# Sparse Representation and Non-Negative Matrix Factorization for image denoise

R. M. Farouk, M. E. Abd El-aziz, A. M. Adam, Zagazig University, Fac. Of Science, Math. Dept.

**Abstract**

Recently, the problem of blind image separation has been widely investigated, especially the medical image denoise which is the main step in medical diagnosis. Removing the noise without affecting relevant features of the image is the main goal. Sparse decomposition over redundant dictionaries become of the most used approaches to solve this problem. NMF codes naturally favor sparse, parts-based representations. In sparse representation, signals represented as a linear combination of a redundant dictionary atoms. In this paper, we propose an algorithm based on sparse representation over the redundant dictionary and Non-Negative Matrix Factorization (N-NMF). The algorithm initializes a dictionary based on training samples constructed from noised image, then it searches for the best representation for the source by using the approximate matching pursuit (AMP). The proposed N-NMF gives a better reconstruction of an image from denoised one. We have compared our numerical results with different image denoising techniques and we have found the performance of the proposed technique is promising.

## INTRODUCTION

In recent years the blind image had more attention, it's considered an advanced image processing technique and has many applications. Since Images always contaminated with noise in the process of image acquisition and transmission phases, denoising became an essential step to improve the image quality. And it is considered as a preprocessing step in many applications such as image segmentation, feature extraction, and medical image applications [1]. So, it is important to remove the noise without affecting the important features of image as much as possible. There are many types of denoising depend on different criteria, such as smoothing filters like adaptive Wiener filter [2], the frequency domain denoising methods [3] to the recently developed methods which uses directional transformations and multiscale like wavelet, ridgelet and curvelet [4], images tend to become sparse in the wavelet domain, in which the image can be represented by using a small subset of the wavelet coefficients, and this is why the wavelet-based models widely used[5-6]. When representing an image with a rich number of local features, these features can't be well represented by only one fixed dictionary [7], and some artifacts will appear in the denoised image. To solve this problem, researchers developed the dictionary learning methods to learn the dictionary from the data instead of using a fixed dictionary. Sparse redundant representation and K-SVD based denoising algorithm were proposed to train a highly over-complete dictionary [8]. shape-adaptive discrete cosine transform (DCT) to the neighborhood [9], This algorithm can achieve very sparse representation of the image and hence lead to effective denoising. Non-negative matrix factorization is different from other methods; it adds its non-negative constraints. When applied to im-age representation, the obtained NMF basis are localized features that correspond with intuitive notions of the parts of the image. some other techniques used Some factorization methods like PCA and SVD to do block wise analysis in order to conduct image denoising by modeling each pixel and its neighborhood as a vector variable [10]. In this paper, to improve the performance of learning parts of images we combine NMF with log-Gabor wavelets. We compare the new method with other methods like Gabor wavelet and DCT to verify its good performance in image denoising.

Non-Negative Matrix Factorization for the dictionary atoms update, instead of using the SVD decomposition for dictionary atoms update and the Orthogonal Matching Pursuit (OMP) for sparse representation for the data matrix. Also, instead of the DCT dictionary used on the K-SVD, we choose the log-Gabor wavelet dictionary as an initial dictionary.

## 1. Sparse representation

Sparse representations of images become one of the most important topics recently, it tends to find a set of atoms $y_i \in R^n$ that form an over-complete dictionary $D \in R^{n \times K}$ that contains K atoms $\{d_i\}_{i=1}^K$ $(K > n)$. The formulation of $y$ may be exact $y = Ds$ or approximate, $y \approx Ds$, satisfying $\|y - Ds\|_p < \varepsilon$, where the vector $s$ is the sparse representation for the vector $x$.

To find $s$ we need to solve either

$(P_0) \min_s \|s\|_0$ subject to $y = Ds$ (1)

Or

$(P_{0,\varepsilon}) \min_s \|s\|_0$ subject to $\|y - Ds\|_2 < \varepsilon$ (2)

where $\| \|_0$ is the $l0$ norm is defined as the number of non-zero elements.

Now suppose that we want to estimate the source signal $z$ from the observed noised version

$y = z + n$, where n is a white Gaussian noise.

Assume $z$ is the source signal and it has a sparse representation over an over-complete dictionary $D$, i.e. $z = Ds$,

where $s$ is the sparse representation of $z$ over the dictionary D.

We can form the problem as in equation (2).

In this paper, we have a new algorithm for solving this problem. Our algorithm much likes the K-SVD algorithm but, we use the Approximate Matching Pursuit for sparse representation for the data matrix and the

### 2.1 Approximate matching pursuit

Given a signal $y \in R^M$, and a dictionary, $D \in R^{M \times K}$, we need to find a vector of coefficients $s \in R^K$ that minimizes $\|y - Ds\|_2$. Orthogonal matching pursuit (OMP) is an iterative algorithm that selects the atom which is most correlated with the current residuals at each step. The approximate matching pursuit (AMP) algorithm described in Algorithm1 is similar to the orthogonal matching pursuit algorithm (OMP), except that, the (AMP) addresses the main computational bottleneck for large dictionaries by using nearest neighbor search by allowing any adequately near neighbor to be selected as a component instead of compute a large amount of inner product.

| **Algorithm 1. Approximate Matching Pursuit (AMP)** |
|---|
| Input: dictionary $D = [d_1, d_2, \ldots, d_K] \in R^{M \times K}$, data $y \in R^M$ |
| Initialization: Let $r = y, s = 0, L = \emptyset, err = y'y$ <br> While $err > \varepsilon$ do |
| Find any i such that $d_i$ and $r$ are Near Neighbors |
| $$L = L \cup i$$ |
| Solve $s_i = \arg\min_{s_{i,i \in L}} \|y - \sum d_i s_i\|$ $$r = y - \sum_{i \in L} d_i s_i$$ |
| End while <br> Output s |

R. M. Farouk, M. E. Abd El-aziz, A. M. Adam. **Sparse Representation and Non-Negative Matrix Factorization for image denoise**## 2. Non-negative matrix factorization (N-NMF)

Factorization the data into simple, fundamental factors make it possible to identify the most meaningful components of data. Many real-world data are nonnegative and the corresponding hidden components have a physical meaning only when non-negative. In practice, both nonnegative and sparse decompositions of data are often either desirable or necessary when the underlying components have a physical interpretation. For example, in image denoising, involved variables and parameters may correspond to pixels, and non-negative sparse decomposition is related to the extraction of relevant parts from the images [11-12].

The basic NMF problem can be stated as follows:

Given a nonnegative data matrix $Y \in R_+^{I \times T}$ with $y_{it} \geq 0$ or equivalently $Y \geq 0$ and a reduced rank $J$ ($J \leq \min(I,T)$), find two nonnegative matrices $D = [d_1, d_2, \ldots, d_K] \in R_+^{I \times T}$ and $S = B^T = [b_1, b_2, \ldots, b_j]$ which factorize $Y$ as well as possible, that is $Y = DS + E = AB^T + E$, where the matrix $E \in R^{I \times J}$ represents approximation error.

## 3. Image denoising based on N-NMF

In this section, we introduce our N-NMF, which uses Nearest Neighbor search with the Nonnegative Matrix Factorization for image denoising. We choose the log-Gabor dictionary as an initial dictionary for some reasons. Firstly, choosing log-Gabor functions allows us to avoid some of the artificial block edge effects of the DCT basis, which used in the K-SVD, since log-Gabor functions tend to decay smoothly at the edges. Separability is another nice property since it reduces the computation necessary to perform the matching search [13]. The other reason is that the log-Gabor dictionary gives a recovery rate higher than the DCT dictionary (Schnass and Vandergheynst 2008). We generate an initial dictionary $D$ of size $M \times K$ ($K \gg M$) from the log-Gabor basis functions. For a noised image of size $M \times M$ we generate blocks of size $\sqrt{M} \times \sqrt{M}$ by using a sliding window moving over the image, and then each block is used as a column of the data matrix Y, which used after that for learning the dictionary. Then we use the N-NMF algorithms, which alternate between the sparse representation of the data with fixed dictionary and updating the dictionary with fixed representation to get the best dictionary to represent the important component in the image. At the end, we use the learned dictionary to reconstruct the source image.

**Algorithm 2. N-NMF algorithm.**

1. Initialization: Set an over-complete dictionary $D = [d_1, d_2, \ldots, d_K] \in R^{M \times K}$, from the log-Gabor wavelet basis (LGW) s.t. $\|d_K\| = 1$ for all k.
2. Repeat until met the error goal
   * Sparse coding: find the sparse representation $S = [s_1, s_2, \ldots, s_N]$ for data matrix $Y = [y_1, y_2, \ldots, y_N]$ based on the fixed dictionary A by using the approximate matching pursuit algorithm (AMP).
   * For each column $i = 1, 2, \ldots, N$ solve $\hat{s}_i = \arg\min_{s_i}\|s_i\|$ s.t. $\|y_i - Ds_i\| \leq \varepsilon$
   * Dictionary update: update the dictionary atoms while fixing the data matrix $Y$ and the sparse representation $s$ by using the Nonnegative Matrix Factorization (NMF) for the overall representation error.
3. Reconstruction: reconstruct the denoised image $I_d = D\hat{s}$

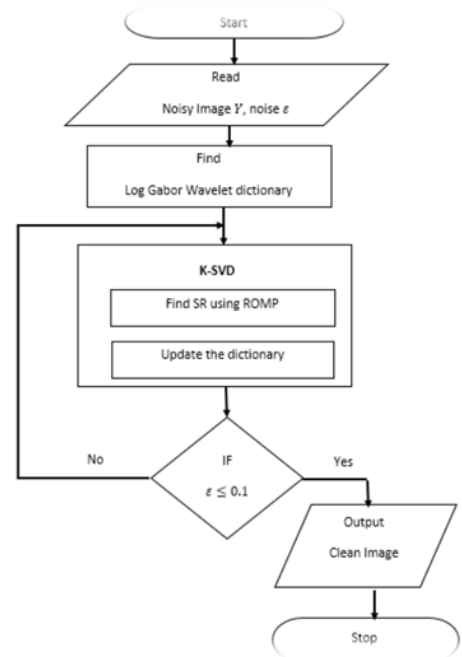

*Figure 1 shows the algorithm flowchart*

R. M. Farouk, M. E. Abd El-aziz, A. M. Adam. **Sparse Representation and Non-Negative Matrix Factorization for image denoise**## 4. Numerical results

To evaluate the performance of our algorithm, it was compared with other algorithms such as KSVD based on Gabor wavelet, DCT. All experiments are performed on a PC running Windows 10 64bit and 4G RAM. The experiments are based on some benchmark MRI images data set from [14], and some Natural Images. The performance of the proposed algorithm was quantified across different noise levels. For each noise level, the average performance was calculated for each algorithm over many runs. To measure the performance of algorithms the Peak Signal to Noise Ratio (PSNR) was used and it was defined as:

$$PSNR = 20\log_{10}(\frac{255}{MSE}) \qquad (3)$$

The results are illustrated in tables 1-2 and figures 2-5. In tables 1-2 the K-SVD using a different type of dictionaries and our algorithm are performed. From this table, the log Gabor wavelet shows very good results, where the original images and the image corrupted by white Gaussian noise are shown in Figures 2-5 (a) and Figures 2-5 (b) respectively. The denoising results obtained by DCT, KSVD, Gabor wavelet, and our algorithm are illustrated in Figures 2-5 (c)-(f) respectively.

## 5. Experimental Results

In this work, we used an over-complete log-Gabor dictionary as an initial dictionary of size 64x256 generated by using log-Gabor filter basis of size 8x8, each basis was arranged as an atom in the dictionary. The dictionary was learned by alternating between sparse coding with the current dictionary and dictionary update with the current sparse representation. For doing that, we use the N-NMF algorithm and the approximate matching pursuit. We applied the algorithm to a medical image (as in Figure. 2 and Figure. 3), Barbra image Lena image (as in Figure. 4 and Figure. 5), from the ORL database of faces (first face in the s1 set as shown in Figure. 4) [15]. The results showed that using the over-complete log-Gabor dictionary with the nonnegative matrix factorization to learn dictionaries for sparse representation gave good results. We used that method for image denoising and evaluate our method by calculating the PSNR and compare our results with the K-SVD methods, which showed that our method gave better results over the K-SVD especially with low-level noise energy. Also, the approximate matching pursuit gave a fast computation compared to the orthogonal matching pursuit used in the K-SVD algorithm.

*Table 1 The PSNR computed for many images with different noise variance level (Sigma).*

| PSNR | DCT | K-SVD | Gabor | Log-Gabor |
|---|---|---|---|---|
| Sigma = 10 | 37.9055 | 38.4330 | 37.8763 | 37.9120 |
| Sigma = 15 | 35.6643 | 36.2256 | 35.5415 | 35.5355 |
| Sigma = 20 | 34.0441 | 34.5714 | 33.9716 | 33.9361 |
| Sigma = 25 | 32.5956 | 33.0668 | 32.4038 | 32.4399 |
| Sigma = 30 | 31.5693 | 31.9675 | 31.3255 | 31.3273 |
| Sigma = 40 | 29.7859 | 29.9928 | 29.5130 | 29.5411 |
| Sigma = 50 | 28.2951 | 28.3773 | 27.9602 | 27.9651 |
| Sigma = 60 | 27.5186 | 27.5778 | 27.2431 | 27.2372 |
| Sigma = 70 | 26.6802 | 26.6282 | 26.3537 | 26.3321 |
| Sigma = 80 | 25.8029 | 25.7539 | 25.5619 | 25.5493 |
| Sigma = 90 | 25.4580 | 25.3633 | 25.1660 | 25.1538 |

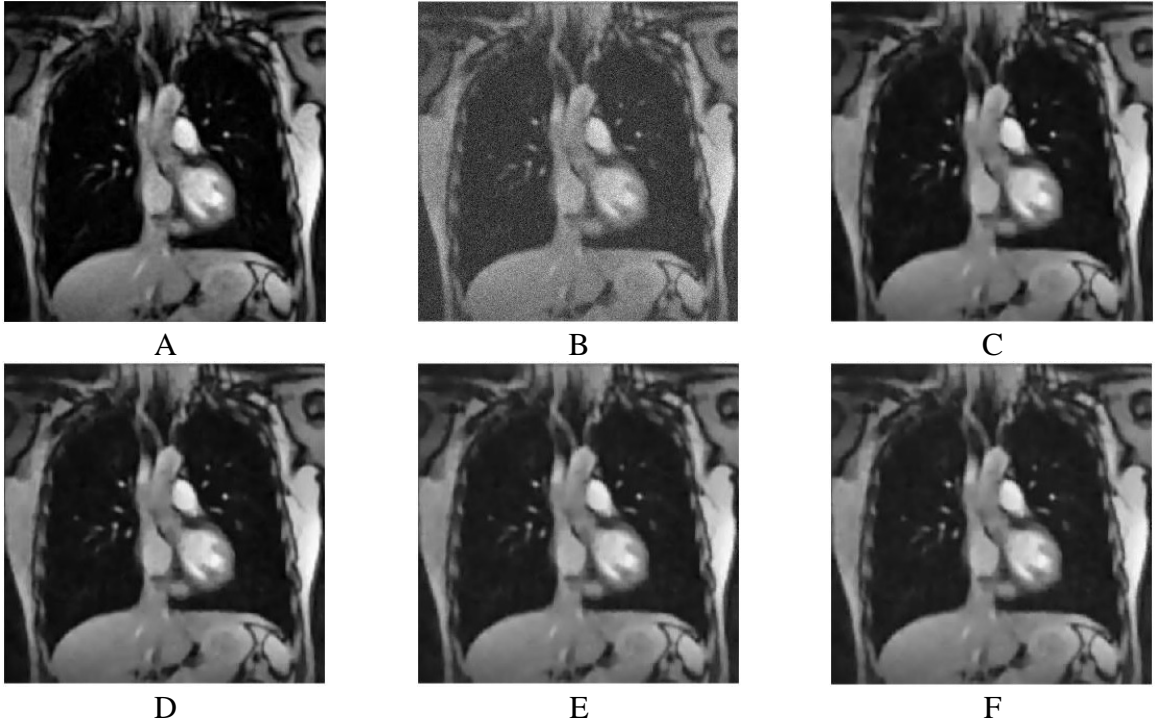

*Figure 2* (a) The original image. (b) The noised image by adding Gaussian noise with sigma=20.
(c) The denoised image by using DCT algorithm and (d) the denoised image by using K-SVD.
(E) The denoised image by using algorithm Gabor wavelet and (F) the denoised image by using N-NMF

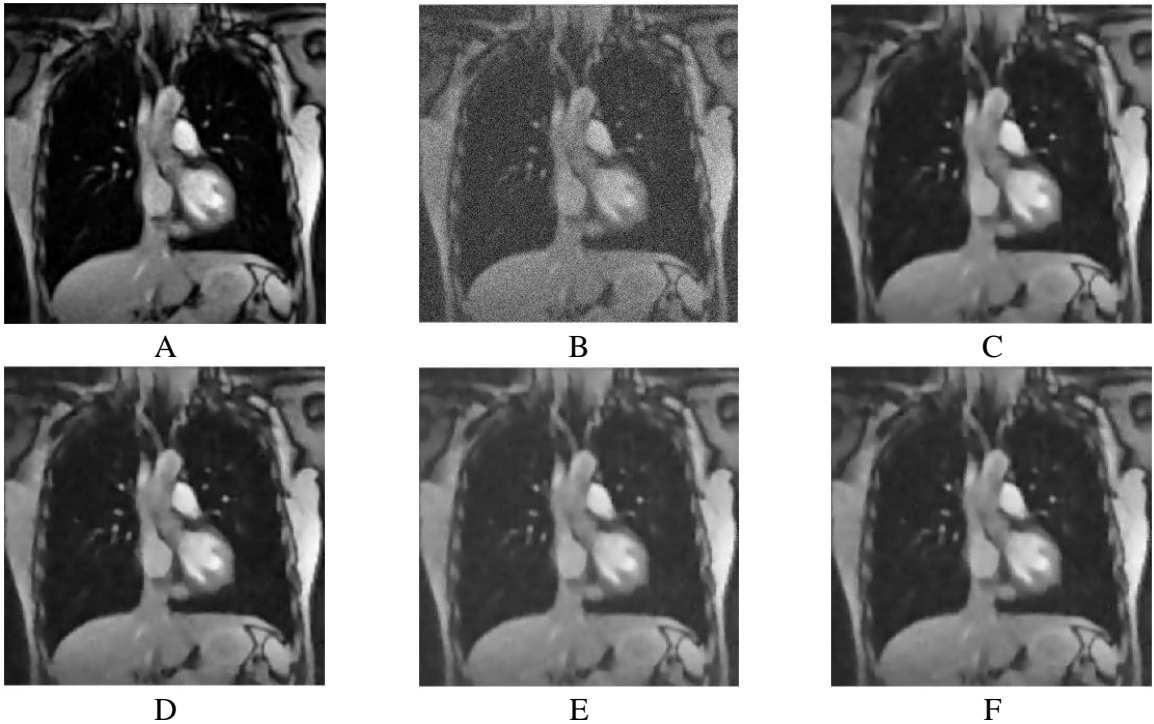

*Figure 3* (a) The original image. (b) The noised image by adding Gaussian noise with sigma=30.
(c) The denoised image by using DCT algorithm and (d) the denoised image by using K-SVD.
(E) The denoised image by using algorithm Gabor wavelet and (F) the denoised image by using N-NMF

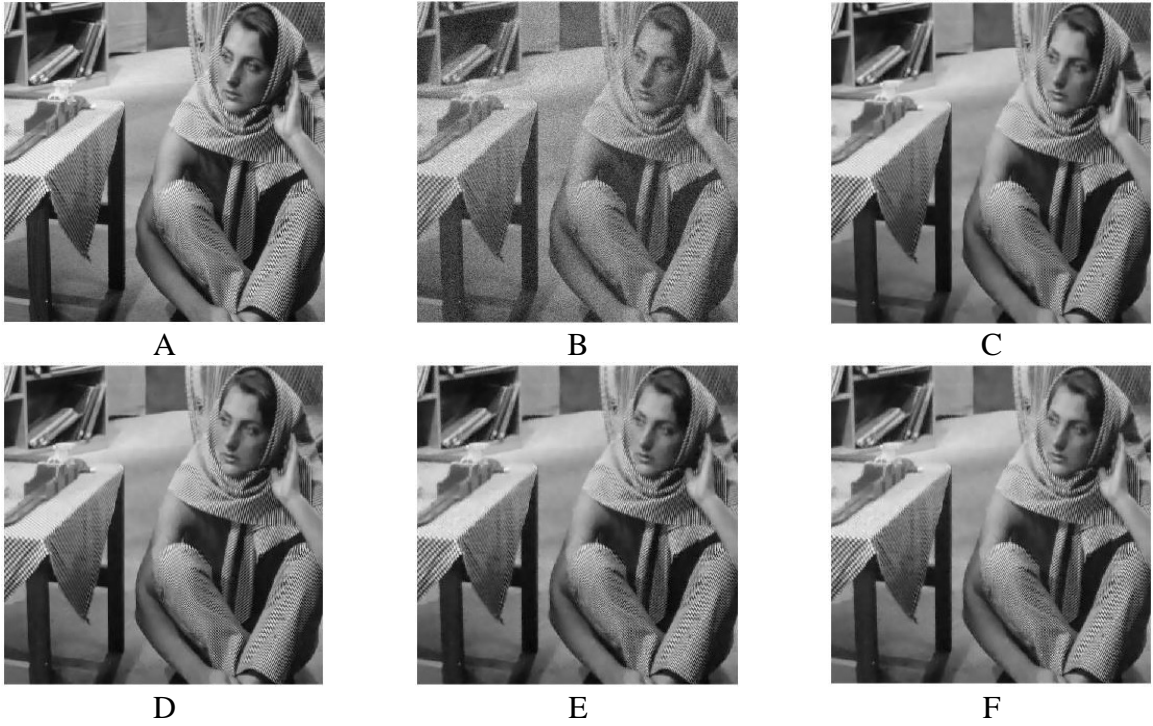

*Figure 4  (a) The original image. (b) The noised image by adding Gaussian noise with sigma=15.
(c) The denoised image by using DCT algorithm and (d) the denoised image by using K-SVD.
(E) The denoised image by using algorithm Gabor wavelet and (F) the denoised image by using N-NMF*

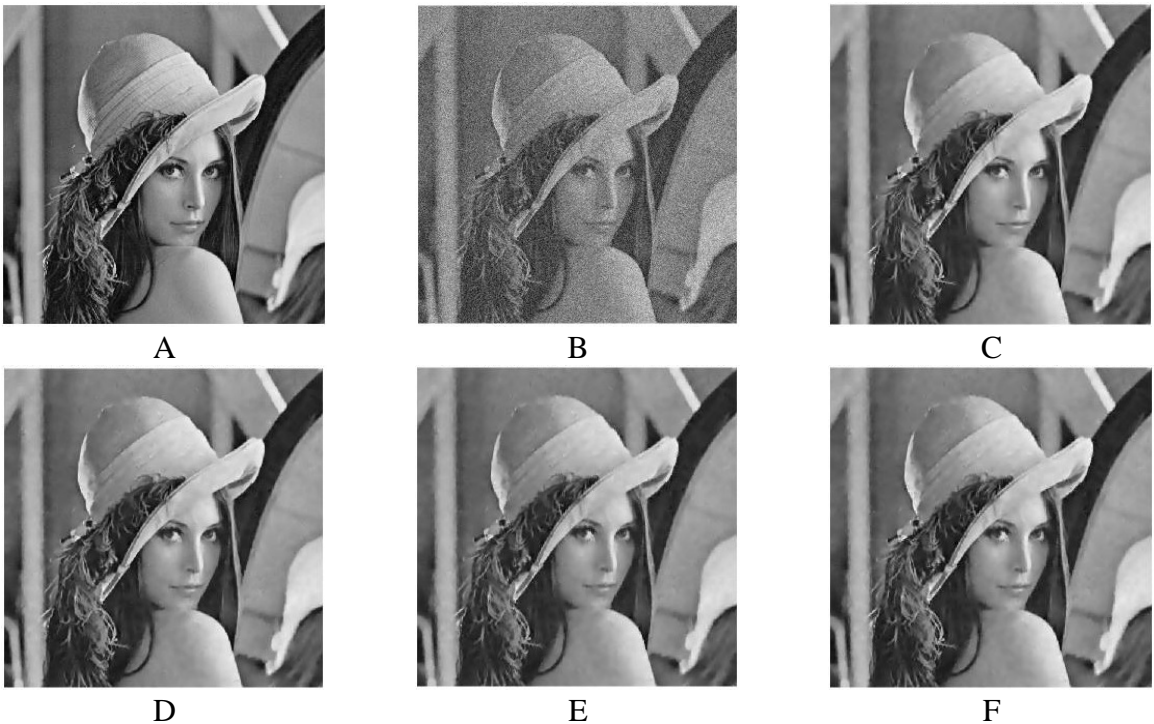

*Figure 5  (a) The original image. (b) The noised image by adding Gaussian noise with sigma=25.
(c) The denoised image by using DCT algorithm and (d) the denoised image by using K-SVD.*

*(E) The denoised image by using algorithm Gabor wavelet and (F) the denoised image by using N-NMF*

| PSNR | | DCT | K-SVD | Gabor | Log-Gabor |
|---|---|---|---|---|---|
| Sigma = 15 | (Barbara) | 31.6451 | 32.4082 | 30.6316 | 30.7359 |
| | (Lena) | 33.4024 | 33.7369 | 33.3922 | 33.3637 |
| Sigma = 20 | (Barbara) | 29.9527 | 30.8145 | 29.2111 | 29.1106 |
| | (Lena) | 31.9886 | 32.4618 | 32.0571 | 32.0253 |
| Sigma = 25 | (Barbara) | 28.6046 | 29.5540 | 27.7520 | 27.8415 |
| | (Lena) | 30.8926 | 31.3833 | 30.9046 | 30.9480 |
| Sigma = 30 | (Barbara) | 27.5846 | 28.5568 | 27.0104 | 26.9122 |
| | (Lena) | 29.9652 | 30.4453 | 30.0192 | 30.0304 |
| Sigma = 40 | (Barbara) | 25.9533 | 26.8214 | 25.5091 | 25.5054 |
| | (Lena) | 28.5516 | 29.0562 | 28.6219 | 28.6253 |
| Sigma = 50 | (Barbara) | 24.8023 | 25.5081 | 24.4021 | 24.4170 |
| | (Lena) | 27.5584 | 27.9332 | 27.5474 | 27.5428 |

## 6. Discussion and conclusion

In this paper, we address the image denoising problem based on sparse coding over an over-complete dictionary. Based on the fact that both nonnegative and sparse decompositions of data are often either desirable or necessary when the underlying components have a physical interpretation, which implies on real images. We presented an algorithm N-NMF, which used the technique of learning the dictionary to be suitable for representing the important component in the image by using the nonnegative matrix factorization technique for updating the dictionary in the learning process and using approximate matching pursuit algorithm for finding the sparse coding of the data based on the current dictionary. Experimental results show satisfactory recovering of the source image. Future theoretical work on the general behavior of this algorithm is on our further research agenda.